\pdfoutput=1
\documentclass[11pt]{article}
\usepackage[final]{acl}
\usepackage{newtxtext}
\usepackage{newtxmath}
\usepackage{latexsym}
\usepackage[T1]{fontenc}
\usepackage[utf8]{inputenc}
\usepackage{microtype}
\usepackage{inconsolata}
\usepackage{graphicx}
\usepackage{booktabs}
\usepackage{verbatim}
\usepackage{multirow}
\usepackage{url}
\usepackage{dblfloatfix}
\usepackage{tabularx}

\usepackage[most]{tcolorbox}
\usepackage{lipsum}

\tcbset{
  compact/.style={
    boxrule=0.4pt,
    left=2pt, right=2pt, top=2pt, bottom=2pt,
    fontupper=\footnotesize,
    sharp corners,
    boxsep=0pt,
  }
}

\definecolor{childes}{HTML}{e41a1c}
\definecolor{babylm}{HTML}{377eb8}
\definecolor{fineweb}{HTML}{4daf4a}
\definecolor{fineweb-short}{HTML}{984ea3}
\definecolor{tinydialogues}{HTML}{ff7f00}

\title{Child-directed speech facilitates production, not comprehension, in BabyLMs}

\author{Bastian Bunzeck \and Sina Zarrieß \\
  Computational Linguistics, Department of Linguistics \\
  Bielefeld University, Germany \\
  \texttt{\{bastian.bunzeck, sina.zarriess\}@uni-bielefeld.de}}

\begin{document}

\maketitle
\begin{abstract}
Recent studies suggest that child-directed speech is not conducive to language learning in BabyLMs. However, current evaluations focus predominantly on comprehension and not production, which is central to usage-based theories of language acquisition which argue how CDS facilitates early language use through constructional ``frames'' (frequent lexical patterns with open slots). We introduce a novel generation-based evaluation inspired by such theories in form of a \textbf{frame-completion task}, and compare Llama models trained with CDS, the BabyLM corpus, and web-crawl data (FineWeb-edu) on comprehension benchmarks and our novel framework. Our results reveal a clear dissociation between models' comprehension and production capabilities: while FineWeb-trained models excel at minimal pairs, CDS-trained models produce grammatical completions substantially earlier in training and concentrate probability mass on appropriate slot-fillers. These findings show that comprehension benchmarks underestimate what CDS affords to BabyLMs.\footnote{Models, prompts and other data can be found \href{https://huggingface.co/collections/bbunzeck/child-directed-speech-facilitates-production-conll-2026}{in this HuggingFace collection}.}
\end{abstract}

\section{Introduction}
Child-directed speech (CDS) differs from regular speech, \textit{inter alia}, through short utterances, exaggerated prosody and frequent repetition, properties that facilitate language acquisition by grabbing children's attention \cite{snow1977talking,fernald1985fourmonthold,soderstrom2007babytalk}.  Usage-based approaches to language acquisition argue that these distinctive distributional properties are significant for human learners, as they provide constructional frames which enable children to extract \textit{productive} patterns for their own usage through item-based learning \cite{tomasello2000itembased,diessel2013construction,behrens2021constructivist} and frequency-driven mechanisms \cite{behrens2009usagebased,diessel2016frequencya}. This enables children to communicate their intentions effectively \cite{raz2020learning}, long before they have a full-fledged, adult-like grammatical system \cite{tomasello2000young}. 

\begin{figure}[t]
\centering
\begin{tcolorbox}[
  compact,
  colback=white,
  width=0.9\columnwidth,
  colframe=black,
  title={\footnotesize\bfseries Frame: \texttt{I like to play with my \_}},
  fonttitle=\footnotesize\bfseries,
  coltitle=black,
  colbacktitle=black!8,
  toptitle=2pt, bottomtitle=2pt,
  top=5pt, bottom=4pt,
  left=5pt, right=5pt,
]
Completion by model trained on data from:
\begin{tcolorbox}[compact, colback=childes!8, colframe=childes, title={\footnotesize \textsc{Childes}}, toptitle=1pt, bottomtitle=0.5pt, coltitle=white, colbacktitle=childes]
\texttt{$\triangleright$ toys.}
\end{tcolorbox}%\vspace{0.5pt}

\begin{tcolorbox}[compact, colback=babylm!8, colframe=babylm, title={\footnotesize \textsc{BabyLM}}, toptitle=1pt, bottomtitle=0.5pt, coltitle=white, colbacktitle=babylm]
\texttt{$\triangleright$ own of the time.}
\end{tcolorbox}%\vspace{0.5pt}

\begin{tcolorbox}[compact, colback=fineweb!8, colframe=fineweb, title={\footnotesize \textsc{FineWeb-edu}}, toptitle=1pt, bottomtitle=0.5pt, coltitle=white, colbacktitle=fineweb]
\texttt{$\triangleright$ be a few of the world.}
\end{tcolorbox}%\vspace{0.5pt}

\begin{tcolorbox}[compact, colback=fineweb-short!8, colframe=fineweb-short, title={\footnotesize \textsc{FineWeb-edu-short}}, toptitle=1pt, bottomtitle=0.5pt, coltitle=white, colbacktitle=fineweb-short]
\texttt{$\triangleright$ .}
\end{tcolorbox}%\vspace{0.5pt}

\begin{tcolorbox}[compact, colback=tinydialogues!8, colframe=tinydialogues, title={\footnotesize \textsc{TinyDialogues}}, toptitle=1pt, bottomtitle=0.5pt, coltitle=white, colbacktitle=tinydialogues]
\texttt{$\triangleright$ toys.}
\end{tcolorbox}

\end{tcolorbox}
\vspace{-0.2cm} 
\caption{Lexical frame completions generated after 10\% of training under greedy decoding, illustrating models' early production capabilities. More examples are presented in Appendix \ref{sec:more-examples}.}
\label{fig:example}
\vspace{-0.6cm} 
\end{figure}

Recently, work on LMs trained with acquisition-scale data (BabyLMs, \citealp{warstadt2023findings}) has focused on the linguistic capabilities acquirable from CDS in neural learners, for example, by systematically manipulating the presence and amount of CDS in pretraining (e.g., \citealp{huebner2021babyberta,padovani2025childdirected,bunzeck2025construction}). Yet, the effect of CDS on language models remains contested. Although a few studies show clear positive effects of CDS on LMs' linguistic capabilities, like faster learning of syntactic specific phenomena \cite{huebner2021babyberta,salhan2024less}, others show no advantage for CDS compared to wiki text \cite{padovani2025childdirected} or a general inability to learn specific hierarchical rules \cite{yedetore2023how}. In practice, however, current BabyLM evaluation resources are limited to tests of implicit grammatical knowledge, e.g., measured with minimal pair datasets \cite{warstadt2020blimp,jumelet2026multiblimp}.
While informative about rule-based knowledge, these evaluations do not assess other aspects of linguistic knowledge that usage-based approaches are interested in (cf. \citealp{weissweiler2025linguistic}), for example if models can produce acceptable and appropriate utterances. Further, while some work has mapped out the learning trajectories of comprehension-based benchmarks (cf. \citealp{choshen2022grammarlearning,bunzeck2024fifty,padovani2025childdirected}), almost nothing is known about how generation capabilities evolve during pretraining. Do models only produce word salad in their early checkpoints (cf. Figure \ref{fig:example}), or do they generate utterances that are simple, short, and largely correct, as in children's speech? This distinction matters for usage-based and constructionist approaches, which predict that CDS facilitates the production of simple but appropriate early language use. Therefore, we hypothesize that the limited effects of CDS in LM training can be attributed to the restricted evaluation paradigm that is not tailored to capture conducive effects of CDS on \textbf{early} language use, which, however, is the central point in usage-based theories \cite{tomasello2003constructing,rowland2025constructing}.

To address this gap, we introduce our frame-completion task, a generation-based evaluation that measures productive capabilities. We pretrain models on a variety of more/less developmentally plausible datasets, prompt them to complete sentence fragments (corresponding to constructional frames), and analyze i) whether completions are grammatical, ii) how certain models are when completing frequent lexical frames, and iii) how generation capabilities develop during pretraining. Our findings support the assumptions made in usage-based linguistics: While models trained on complex corpora like \textsc{FineWeb-edu}\footnote{We designate datasets in small caps (e.g., \textsc{FineWeb-edu}) and models trained on them in monospace (\texttt{FineWeb-edu}).} \cite{penedo2025fineweb2} surpass models trained on CDS on minimal pair benchmarks, CDS-trained models produce more grammatical frame completions earlier in training. Furthermore, entropy analyses show that CDS models make more focused predictions, e.g., concentrating probability mass on semantically appropriate concrete nouns in argument positions. This pattern suggests CDS-trained models learn constructional templates that scaffold productive combination and the correct completion of lexical frames with appropriate lexical items early in pretraining, whereas models trained on web text attempt to generate overly complex syntactic structures. This divergence is hidden when \textit{only} using minimal pairs to measure the linguistic knowledge of BabyLMs.

\section{Related work}

\paragraph{Usage-based language acquisition}

A central tenet of usage-based language acquisition is that linguistic knowledge is built up through \textit{use} in an active learning process \cite{tomasello2000first,raz2020learning,rowland2025constructing}. Using language means achieving communicative goals through its production \cite{diessel2017usagebased}. Children do so long before their language system is fully abstract. Production then is not merely a reflection of acquired knowledge, but a driver of it. In the preverbal stage, children use grunts, babbling and pointing to achieve communicative goals \cite{mccune2008how,mcgillion2017what}, which already improves motor skills necessary for production, before moving on to semantically motivated isolated words (e.g., \textit{there}, \textit{mommy}) and holistic phrases (e.g., \textit{get-it}, \textit{all-gone}). Early multiword utterances typically revolve around a `pivot word' (a fixed anchor like \textit{More \_} or \textit{Want \_}) with an open slot (cf. \citealp{braine1976children,lieven1997lexicallybased,tomasello2000itembased}) and serve particular speech-act functions \cite{tomasello1992first,tomasello2003constructing}. These utterances are reflected in the input: The vast majority of CDS utterances combine a highly frequent \textit{frame} for the utterance with an open slot. For example, copular clauses are typically introduced by a pronoun and an auxiliary followed by a slot for nominals (e.g., \textit{That’s}/\textit{It’s} + ENTITY). Such frames lay the ground for children’s early utterances (cf. Section \ref{sec:frames}).

\begin{table*}[ht!]
\centering
%\small
\begin{tabular}{@{}lr|lr@{}}
\toprule
Child speech frames & Freq. & FineWeb-edu frames  & Freq. \\ \midrule
\textit{I like to play with my} \_& 22 & \textit{It is one of the most} \_ & 122\\
\textit{and then you put it in} \_& 22 & \textit{It is interesting to note that} \_ & 106\\
\textit{but I don't know how to} \_ & 21 & \textit{All you have to do is} \_ & 92 \\
\textit{when I grow up I want} \_& 19 & \textit{The reason for this is that} \_ & 91 \\
\textit{I don't know how to get} \_& 17 & \textit{This is due to the fact} \_ & 91 \\
\textit{but I don't know where the} \_& 17 & \textit{Note that depending on the number} \_ & 90 \\
\textit{and then put it in the} \_& 17 & \textit{It can also be used to} \_ & 84 \\
\textit{and then they went to the} \_& 16 & \textit{This is one of the most} \_ & 77\\
\textit{and how I would do it} \_& 16 & \textit{It should also be noted that} \_ & 75\\
\textit{can I have a bit of} \_& 15 & \textit{It is a good idea to} \_ & 72\\
 \bottomrule
\end{tabular}
\caption{Six-word frames used for prompting, including frequency in the dataset. For full list of frames see Table \ref{tab:all-frames}.}
\vspace{-0.3cm} 
\label{tab:six-word-frames}
\end{table*}

\paragraph{CDS in BabyLMs}
\label{sec:cds-babylm}

Recent BabyLM studies have produced contradictory findings on the effect of purely CDS-based pretraining on linguistic knowledge. On the positive side, CDS has been shown to improve performance on various benchmarks (Zorro, BLiMP, cloze tests) over traditional pretraining data such as Wikipedia in masked and autoregressive LMs \cite{huebner2021babyberta,qin2024systematic,feng2024childdirected}, across languages \cite{salhan2024less}, as pretraining for further fine-tuning \cite{mueller2023how}, for in-context learning \cite{deshpande2023honey,muckatira2024emergent} and for completing child-caretaker dialogues in contrast to fine-tuned LLMs \cite{levandovsky2025learning}. Mixed to negative results are found for capturing hierarchical generalizations in question formation \cite{yedetore2023how}, predicting word learning trajectories \cite{ficarra2025distributional}, but also for minimal pair benchmarks when comparing to book text \cite{yam2024what}, when comparing to construction distributions in written and spoken language, where written profiles outperform spoken ones on MP benchmarks \cite{bunzeck2025construction}, when comparing to Wiki data across languages (English, French and German, \citealp{padovani2025childdirected}), to 5-gram models \cite{vazquezmartinez2023evaluating}, or when training on dialogue \cite{padovani2025dialogue}. These contradictions likely stem from differences in evaluation method and comparison baselines, but also from a shared reliance on comprehension-based tasks. This is partly architectural: Many BabyLMs are masked language models unsuitable for text generation, and even autoregressive ones lack the reinforcement-tuning that makes large models fluent generators.

\paragraph{Evaluating production}
Production-based evaluations have so far received little attention. \citet{pannitto2020recurrent} take an explicit usage-based stance and train LSTMs (not Transformer LMs) on 3M words of CHILDES data, OpenSubtitles, and Simple English Wikipedia. Across several checkpoints, they generate text and extract specific subtrees, from which they find that CHILDES-trained models approximate their input best, which they attribute to the repetitiousness of CDS. \citet{nikolaus2021modeling} show that for a language-and-vision task, production-based learning through corrective feedback improves performance over perception-only learning (note, however, that the evaluation method stays the same here). Using a masked language model trained on the BabyLM corpus (CDS + other data sources), \citet{rozner2025babylms} show that such small models have high affinity between open slots and their appropriate fillers, even for abstract constructions such as \textit{let-alone} and \textit{much-less}. \citet{lee2025readability} train autoregressive models on synthetic child stories and adult-oriented texts, and measure completion quality through traditional readability measures (e.g., Flesch-Kincaid score) and LLM-as-a-judge scoring. They find that models trained on supposedly more ``readable'' text generate coherent completions later, not earlier, than models trained on less readable data. In fact, learnability comes from less n-gram diversity, not readability measures. \citet{levandovsky2025learning} train autoregressive models on dialogue from CHILDES and use an LLM-as-a-judge approach to measure completion coherence. Here, a model completely trained on CHILDES outperformed larger pretrained models that were only fine-tuned on CHILDES data, and a robot powered by this model was rated as more child-like by human test subjects. While these generation-based studies show mixed results for training on simple language or CDS, they respond to an emerging call for more production-based evaluation: In particular, \citet{weissweiler2025linguistic} argue that evaluations should use natural stimuli and focus more on partially-filled schemas and constrained slots, exactly because current probing methods like MPs are not sufficient for constructionist approaches to language. Our study contributes to this line of work by introducing generation-based measures and asking if they converge on similar conclusions to comprehension-based measures like minimal pairs.

\section{Lexical frames anchor production}
\label{sec:frames}

\paragraph{Lexical frames in usage-based theory}
Usage-based linguistics generally considers frequent lexical patterns to be the principal input and output of early language acquisition.
The first corpus-driven characterization of these lexical patterns was provided by \citet{cameron-faulkner2003construction}, who investigated what they call ``lexical frames'', highly frequent utterance-initial lexical patterns that introduce different types of elements for which they provide open slots (for example \textit{What is \_?} or \textit{There \_ go/es}). They found that over 65\% of the utterances children hear are introduced by such frames, which they identified based on a study-specific frequency criterion (occurring at least 4 times per recording, used by at least half of the recorded mothers). Moreover, they showed that children's use of item-specific patterns is only loosely correlated with frames in the input, constrained to constructions simple enough for children, and often adapted, as in cases of deictic substitution between first- and second-person pronouns in dialogue. This shows that, while children learn from such patterns in an item-based fashion, they not only regurgitate the input but derive their own formulaic patterns. These results hold across languages and samples (cf. \citealp{stoll2009lexically, arnon2016nature, bunzeck2025richness}).

\paragraph{Frame set for evaluation}
Given the prevalence of such frames in learners' input and output, we identify them as the ideal source of prompts for our generation evaluation. For that, we extract the most frequent utterance beginnings of four corpora: Child speech from CHILDES \cite{macwhinney2000childes} (which we do \textit{not} use for pretraining), CDS from \textsc{Childes} (which we use as pretraining data); one subset of FineWeb-edu \cite{penedo2024fineweb} for target frames, and another subset (\textsc{FineWeb-edu}) that we use for pretraining. We assign frame status to $n$ sentence-initial word sequences that occur at least 10 times in sentences with a length $> n$. All datasets feature 10M tokens, except the child speech from \textsc{Childes}, for which only 7M words are available. We then subtract the frames in the pretraining data from the set of frames in the held-out data. This means that there is \textit{no} overlap in frames between training and evaluation data, and any frame that we use for generation is infrequent in the training data. 
This ensures that we test for frame completions on data that is at most infrequent in the training data, rather than on frames that models may simply memorize. Furthermore, children's own frames diverge from CDS frames, as discussed above and by \citet{cameron-faulkner2003construction}. 
This leaves 11,000 candidate frames for the \textsc{Childes} child speech and 8,000 candidate frames for the held-out \textsc{FineWeb-edu} dataset.

To keep our prompting dataset concise, we select the top 10 most frequent 6-word, 5-word, 4-word, 3-word, and 2-word beginnings from both sets, manually discarding formulaic or boilerplate frames (e.g., Wikipedia boilerplate from page footers, nursery rhymes, or counting/spelling exercises). This results in 100 prompts (50 from \textsc{Childes} and 50 from \textsc{FineWeb-edu}). Table \ref{tab:six-word-frames} displays the 6-word frames (full data in Table \ref{tab:all-frames}, Appendix \ref{sec:all-frames}). For both data sources, the general frame structure is fairly similar: words are generally short, and the subject position is mostly filled by personal or demonstrative pronouns. Yet, there are also differences: The \textsc{Childes} frames include word order typical of questions, whereas the \textsc{FineWeb-edu} frames only introduce propositional sentences, frequently featuring a form of the copula verb \textit{be}.

\section{Pretraining}

\paragraph{Data}
We pretrain our LMs on five different corpora of 10M lexical words (approximating the lexical input a child has received between 2 and 5 years of age, as per \citealp{warstadt2023findings} and \citealp{gilkerson2017mapping}): (i) the regular \textsc{BabyLM} corpus \cite{hu2024findings}, which aims to represent the whole breadth of child-available input, (ii) child-directed speech (\textit{not} child speech) from \textsc{Childes}, which maximizes developmental plausibility by only including data attested in child-caretaker interactions, and (iii) \textsc{FineWeb-edu}, which contains web crawls filtered for educational value and closely resembles the data that standard LLMs are trained on, while maintaining a higher diversity than e.g., commonly-used Wikipedia dumps. As additional comparisons, we also train models on the length-restricted \textsc{FineWeb-edu-short}, featuring sentences with at most six words, and on synthetic child-directed speech from \textsc{TinyDialogues} \cite{feng2024childdirected}. Notably, these corpora differ tremendously in their number of types, ranging from 25,104 in \textsc{TinyDialogues} to 341,476 in \textsc{FineWeb-edu-short} (cf. also Appendix \ref{sec:data-diversity}).

\paragraph{Models}

We train autoregressive BabyLMs as they can be used for probability- \textit{and} generation-based evaluations. We use a Llama architecture modeled after \texttt{SmolLM2-135M} \cite{allal2025smollm2}, which prioritize depth over width \citep{petty2024impact,gupta2025how} and have been shown to perform strongly across a variety of benchmarks when 
compared to LMs of similar size. We train for three epochs and save checkpoints after 1\% and 10\% of pretraining and after every completed epoch, resulting in five checkpoints per model. We name models after their training data (\texttt{CHILDES}, \texttt{BabyLM}, \texttt{FineWeb-edu}, \texttt{FineWeb-edu-short}, and \texttt{TinyDialogues}) and make them available on HuggingFace.

\begin{table*}[htb!]
\centering
%\small
\begin{tabular}{@{}l|cc|rrr@{}}
\toprule
 & \multicolumn{2}{c|}{Acceptable generations {\color{gray} (CIs)}} & \multicolumn{3}{c}{MP benchmarks}  \\ \midrule
Model & CHILDES & FineWeb-edu & BLiMP & ZORRO & MultiBLiMP \\ \midrule
\texttt{CHILDES} & \textbf{92.8\%} {\color{gray} [88.8, 96.7]} & 44.1\% {\color{gray} [36.2, 52.0]} & 63.7\% & 73.8\% & 58.6\% \\
\texttt{BabyLM} & 80.9\% {\color{gray} [74.3, 86.8]} & 37.5\% {\color{gray} [29.6, 45.4]} & 67.4\% & \textbf{81.0\%} & 82.6\% \\
\texttt{FineWeb-edu} & 44.1\% {\color{gray} [36.2, 52.0]}& 21.1\% {\color{gray} [14.5, 27.6]}& \textbf{68.0\%} & 76.6\% & \textbf{91.0\%} \\
\texttt{FineWeb-edu-short} & 59.9\% {\color{gray} [52.0, 67.8]}& 38.2\% {\color{gray} [30.3, 46.1]}& 61.9\% & 68.7\% & 52.9\% \\
\texttt{TinyDialogues} & 91.4\% {\color{gray} [86.8, 95.4]}& \textbf{61.8\%} {\color{gray} [53.9, 69.1]}& 61.3\% & 68.8\% & 57.4\% \\ \bottomrule
\end{tabular}
%\vspace{-0.5em}
\caption{i) Proportions of acceptable frame completions (generated with a temperature of 0.8 and nucleus sampling with $p = 0.9$) with {\color{gray} bootstrapped 95\% confidence intervals}, and ii) accuracies on MP benchmarks for our Llama models trained on \textsc{Childes}, \textsc{BabyLM}, \textsc{FineWeb-edu/-short}, and \textsc{TinyDialogues} after three epochs.}
\vspace{-1em}
\label{tab:general-results}
\end{table*}

\section{Evaluation}
\label{sec:evaluation}

\subsection{Comprehension}
As comprehension-based benchmarks, we use BLiMP \cite{warstadt2020blimp}, which covers a wide array of semanto-syntactic phenomena with synthetic minimal pairs; Zorro \cite{huebner2021babyberta}, which is based on BLiMP but restricted in vocabulary to lexical forms found in CHILDES; and MultiBLiMP \cite{jumelet2026multiblimp}, which contains minimal pairs targeting agreement phenomena that were derived from existing UD-treebanks.

\subsection{Production/frame completion}

\paragraph{Decoding}
To complete our lexical frames, we prompt all 25 models and generate at most 32 new tokens. We use nucleus sampling ($p = 0.9$) with a temperature of 0.8, following  best practices for open-ended generation \cite{holtzman2020curious,zarriess2021decodinga} that balance output diversity  with textual coherence. Given our large prompt set (100 prompts) and number of models, we generate three completions per prompt rather than doing extensive sampling per prompt to prioritize the variety of prompts over depth per individual prompt. Appendix \ref{sec:temp-comparisons} contains results for different sampling strategies and temperature settings. For the sake of evaluation, we consider only the first complete generated sentence (marked by the first occurrence of sentence-final punctuation, which every generated data point contained without exception).

\begin{figure*}[ht!]
\centering
\includegraphics[width=\linewidth]{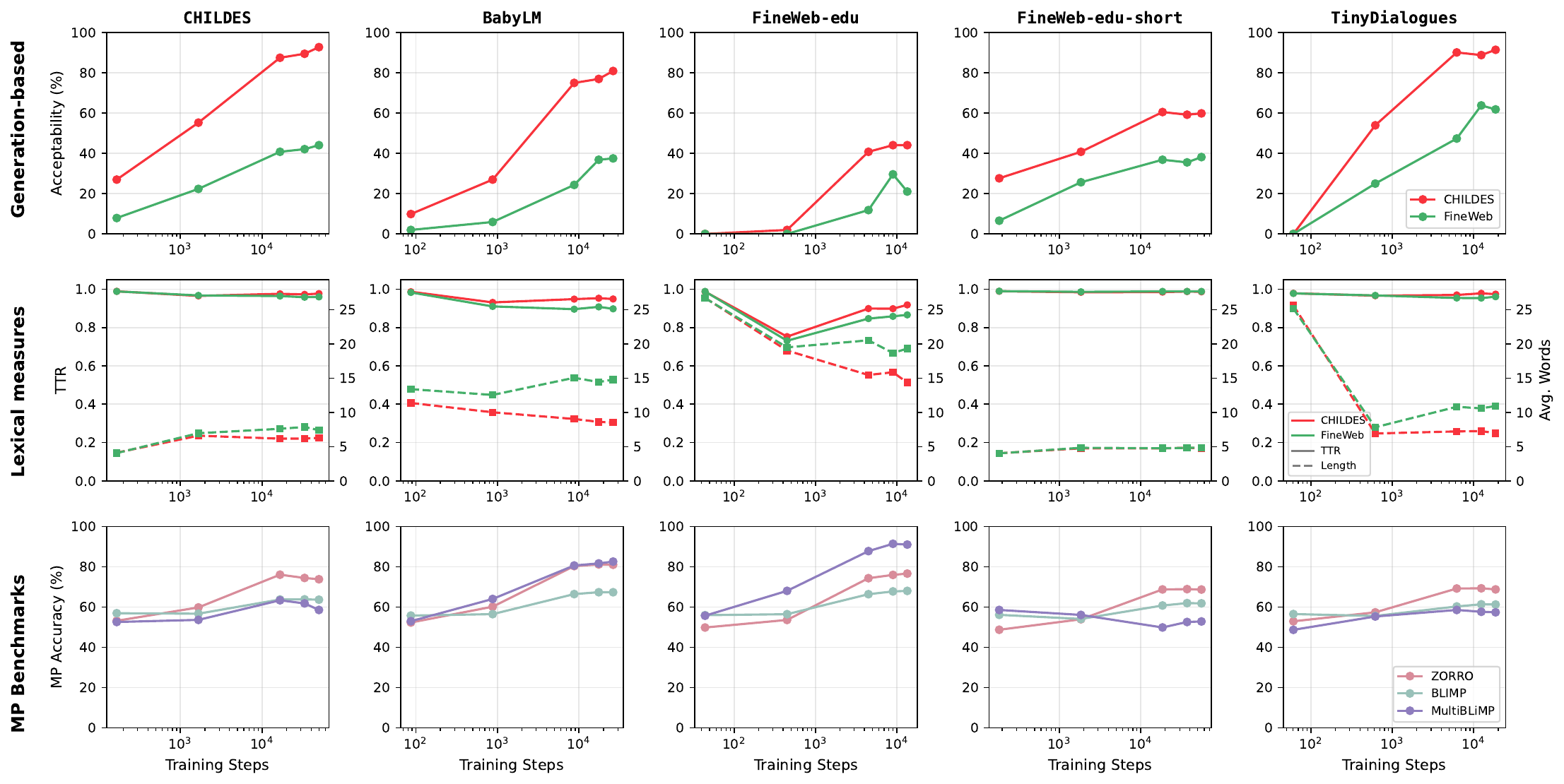}
\vspace{-1.5em} 
\caption{Development of acceptability of generated text, lexical measures, and MP benchmarks.}
\label{fig:main-results}
\vspace{-0.8em}
\end{figure*}

\paragraph{Quality of generations} 
\label{sec:llm-judge-method}
To assess the quality of the generated text, we use an LLM-as-a-judge strategy, following \citet{lee2025readability} and \citet{levandovsky2025learning}. In comparison to other tasks, LLMs are fairly reliable in acceptability ratings \cite{bavaresco2025llms}. In line with best practices \cite{lee2025checkeval}, we frame our evaluation task as a binary decision for acceptability (yes/no). Because no further context is provided for the generated sequences, acceptability here entails being free of syntactic or utterance-internal semantic errors. As the first study of this kind, we restrict our focus to this more simplistic notion of acceptability, although it is also conceivable that reliable judge LMs could grade the generated utterances on further rubrics such as meaningfulness or communicative effectiveness, or even use more complex scoring systems such as magnitude estimation. We use \texttt{qwen3-4b-2507} \cite{yang2025qwen3} as our judge model, as it provided consistent annotations that strongly align with expert judgements on a held-out test set and outperformed comparable models by a wide margin (cf. Appendix \ref{sec:llm-as-judge} for a comprehensive evaluation). We use this model to annotate all generated sequences with acceptability labels. Further, we calculate sequence length and TTR for all sequences as measures of lexical diversity.

\paragraph{Slot-wise productivity} 
We annotate all 100 lexical frames with their ``canonical'' next element (NP, VP, clause, or unclear). To further analyze the development of our models' generative capabilities, we calculate two measures for next-token prediction across our frames: Shannon Entropy $H$ and maximum probability $P_{max}$. 

$H$ quantifies the average uncertainty or ``information content'' in the model's predicted probability distribution. We calculate $H$ as follows, with $V$ being the vocabulary size and $p_i$ being the probability of the $i$-th token:
$H(P) = -\sum_{i=1}^{V} p_i \log_2(p_i)$.

The maximum probability $P_{max}$ is an indicator of whether the model has a clear preference for the next token, calculated as the maximum value in the Softmax output:
$P_{max} = \max_{i \in V} \{p_i\}$.

\section{Results}

Table \ref{tab:general-results} displays evaluation scores for the final models on our novel (generation-based) frame-completion task and established (comprehension-based) minimal pair evaluations. For the production-based measures, we bootstrap confidence intervals by drawing 10,000 resamples of the binary judgments and taking the 2.5th/97.5th percentiles of the resulting means \cite{efron1993introduction}. Wilson 95\% CIs \cite{wilson1927probable} were calculated as a sanity check and yielded near-identical values.

\paragraph{Production-based evaluation with lexical frames}
For text generation prompted with frames from \textsc{Childes} child speech, the \texttt{CHILDES} model trained on child-directed speech is the clear winner, producing 92.8\% acceptable generations (Table \ref{tab:general-results}). The synthetic \texttt{TinyDialogues} model also scores above 90\%, and confidence intervals overlap considerably between these models. The \texttt{BabyLM} model, which also contains CDS from \textsc{Childes}, performs slightly worse at over 80\%, while both web-trained models yield the lowest acceptability rates (44.1\% for \texttt{FineWeb-edu}, 59.9\% for \texttt{FineWeb-edu-short}). This pattern might not seem overly surprising, given that models containing spoken data perform well on spoken-style frames. However, results on \textsc{FineWeb-edu} prompts show a similar pattern: The \texttt{TinyDialogues} model performs best (61.8\%) and \texttt{CHILDES} also fares moderately well (44.1\%). In contrast, the \texttt{FineWeb-edu-short}, with a maximum sentence length of 6, achieves only 38.2\% (although confidence intervals overlap with \texttt{CHILDES} results). The regular \texttt{FineWeb-edu} model performs worst by a clear margin, at 21.1\% (although pretraining data and prompts are drawn from the same underlying dataset). We further report pairwise statistical comparisons in Appendix \ref{sec:pairwise} and provide a qualitative analysis of two three-word frames in Appendix \ref{sec:more-examples}.

\begin{figure*}[hb!]
\centering
\includegraphics[width=\linewidth]{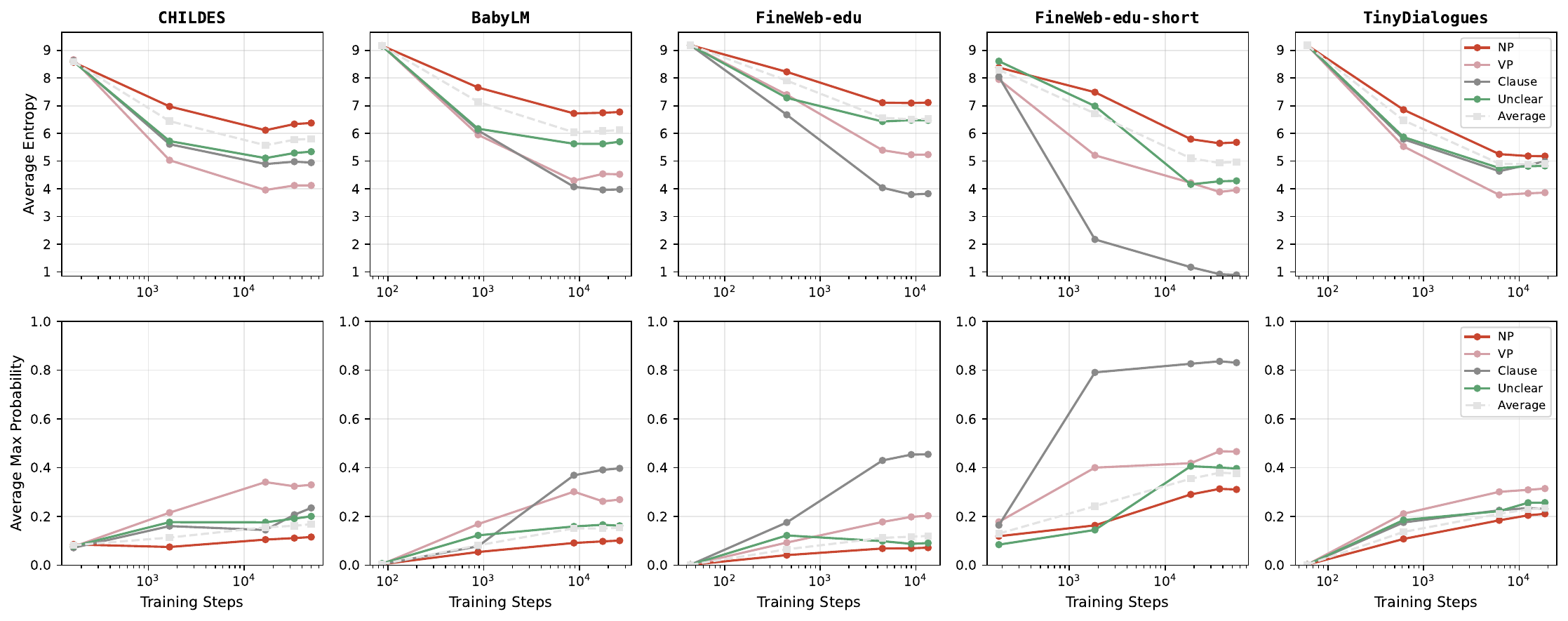}
\vspace{-0.3em}
\caption{Development of slot-wise measures (entropy and max. probability), separated by canonical slot element.}
\label{fig:entropy}
%\vspace{-0.5em}
\end{figure*}

\paragraph{Comprehension-based evaluation}
Performance on the MP benchmarks follows a drastically different pattern (Table \ref{tab:general-results}). While all models score above chance across all MP benchmarks, the \texttt{FineWeb-edu} model, which generates almost no acceptable text, performs best on BLiMP and MultiBLiMP. The \texttt{BabyLM} model achieves the highest score on Zorro. The best-generating \texttt{CHILDES}, on the other hand, only performs reasonably well on Zorro, but poorly on MultiBLiMP. Interestingly, the other strong generation model, \texttt{TinyDialogues}, also dramatically underperforms on all MP benchmarks.

\begin{figure*}[htb!]
\centering
\includegraphics[width=\linewidth]{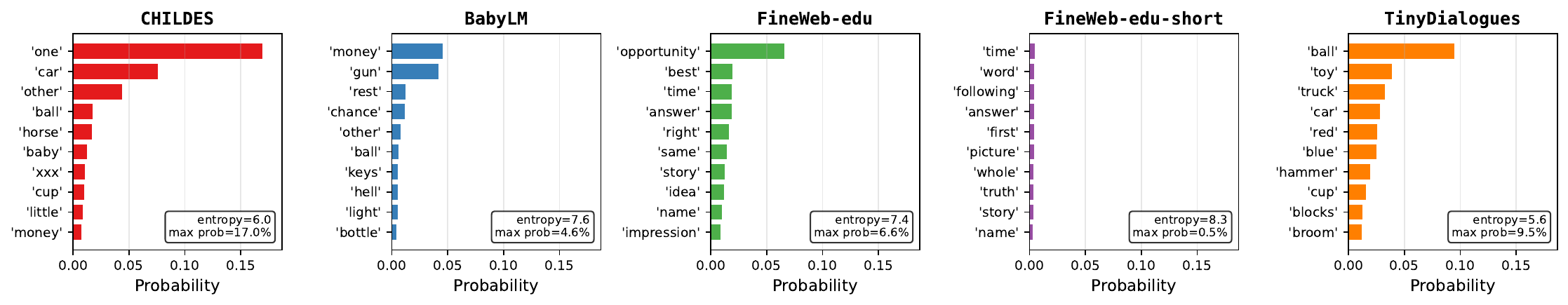}
%\vspace{-1.5em}
\caption{Next token predictions for \textit{Give me the} at final checkpoint (3 epochs). Note that \textit{xxx} is a CHILDES transcription convention for unintelligible speech.}
%\vspace{-1em}
\label{fig:next-token-final}
\end{figure*}

\paragraph{Developmental trajectories} 
Figure \ref{fig:main-results} displays the developmental trajectories for the evaluation measures discussed above. Generally speaking, frame-completion success on \textsc{Childes} and \textsc{FineWeb-edu} prompts improves in tandem across all models. Apart from that, the difference between CDS-trained models and models trained on traditional LM training data is striking. Both natural (\texttt{CHILDES}) and synthetic (\texttt{TinyDialogues}) models exhibit steep improvement curves between 1-10\% of pretraining. In contrast, the \texttt{BabyLM} and \texttt{FineWeb-edu-short} models improve markedly later. The regular \texttt{FineWeb-edu} is the most extreme outlier, as it only starts to generate acceptable utterances after one full epoch of pretraining.

This outlier status is also confirmed by the lexical measures (middle row of Figure \ref{fig:main-results}). Generally, type-token ratio stays close to 1, indicating that models do not generate repeated words, except for \texttt{FineWeb-edu}, which tends to generate repetitive output early in pretraining. The length of generated sequences is more informative: It is low and remains low for \texttt{CHILDES} and \texttt{FineWeb-edu-short}, and somewhat longer for \texttt{BabyLM}. For \texttt{FineWeb-edu} and \texttt{TinyDialogues}, the picture differs substantially: both start out with very long sequences (over 25 words), which then decrease: \texttt{FineWeb-edu} stabilizes between 15-20 words, while \texttt{TinyDialogues} decreases to 5-10 words, comparable to \texttt{CHILDES} (cf. also examples in Figure \ref{fig:example}).

The trajectories of the MP benchmarks are more uniform and start out at a higher level of performance (i.e., due to the nature of the MP scoring task). The largest improvements happen between 10-100\% of pretraining, with the \texttt{BabyLM} and \texttt{FineWeb-edu} models improving the most. Improvements across all three benchmarks happen in tandem; only the \texttt{FineWeb-edu-short} remains an outlier with surprising performance decreases on MultiBLiMP.

\paragraph{Slot-wise entropy and maximum probability}
Figure \ref{fig:entropy} shows the development of Shannon entropy $H$ and $P_{\text{max}}$ for our 100 prompts, separated by the canonical slot fillers. While $H$ generally decreases across training, $P_{\text{max}}$ increases. For the \texttt{CHILDES} and \texttt{TinyDialogues} models, the final checkpoints see a slight increase in entropy without a decrease in $P_{\text{max}}$. The highest entropy ($\approx 7$) is maintained by the \texttt{FineWeb-edu} model, followed by \texttt{CHILDES} and \texttt{BabyLM} ($\approx 6$), while \texttt{FineWeb-edu-short} and \texttt{TinyDialogues} reach the lowest scores ($\approx 5$). Regarding fillers, all models show the highest entropy for NPs. Lower entropies differ between VP (for \texttt{CHILDES} and \texttt{TinyDialogues}) and clause (for \texttt{BabyLM}, \texttt{FineWeb-edu}, and \texttt{FineWeb-edu-short}). Similar trends are visible for $P_{\text{max}}$, where \texttt{FineWeb-edu} reaches the lowest average and \texttt{FineWeb-edu-short} the highest. Again, the VP category is the outlier for \texttt{CHILDES} and \texttt{TinyDialogues}, which assign highest $P_{\text{max}}$ there.

\paragraph{Case study: \textit{Give me the \_}}
Figure \ref{fig:next-token-final} displays the top ten next-word predictions for the lexical frame \textit{Give me the \_} at the final checkpoint of all models (see Appendix \ref{sec:next_token-case_study} for intermediate checkpoints). The canonical element for this slot would be a noun phrase, for which all predicted tokens are valid candidates (mostly nouns, few adjectives). Regarding semantic properties, both \texttt{CHILDES} and \texttt{TinyDialogues} predictions focus on developmentally plausible words like \textit{ball} and \textit{car}. Interestingly, the \texttt{CHILDES} model also distinctively deviates from this pattern: the first and third most probable tokens are the discourse-deictic expressions \textit{one} and \textit{other} that are commonly used for tracking and distinguishing referents, which is a fundamental function of early referential communication \cite{tomasello2003constructing}. No other model predicts such words as particularly likely. In contrast, the \texttt{FineWeb-edu} models predict rather abstract terms (e.g., \textit{time}, \textit{answer}, \textit{story}), and the \texttt{BabyLM} model's predictions appear to be heavily influenced by the unfiltered OpenSubtitles data in the corpus (\textit{money} and \textit{gun} are the most likely predictions).

Concerning slot-level metrics, there is (again) a clear distinction between CDS-trained and remaining models. Models trained on natural/synthetic spoken data are characterized by low entropy and high $P_{\text{max}}$: \texttt{TinyDialogues} has the lowest entropy (5.6) and a $P_{\text{max}}$ of 9.5\%, whereas \texttt{CHILDES} entropy is somewhat higher (6.0), and $P_{\text{max}}$ is 17\%. In these models, probability distributions look most Zipfian \cite{zipf1935psychobiology,baroni2009distributions}, with probability mass being distributed across a small set of tokens with distinctive ``winners''. In contrast, the highest entropy is reported for \texttt{FineWeb-edu-short}, with many tokens being equally (un)likely at 0.5\%. \texttt{BabyLM} and \texttt{FineWeb-edu} show similar high entropies (7.4--7.6) and low $P_{\text{max}}$ (4.6\%--6.6\%). This means the web-trained models have less clear preferences for next tokens.

\section{Discussion}

In general, the results of our experiments confirm our hypothesis. The limited positive effects of CDS observed in prior BabyLM research can  be plausibly attributed to restricted evaluation paradigms. When assessed on production instead of comprehension, CDS-trained models clearly outperform models trained on ``typical'' LM pretraining data. Four findings stand out: i) both \texttt{CHILDES} and \texttt{TinyDialogues} models achieve the highest acceptability rates (90--94\%), both on \textsc{Childes} and on \textsc{FineWeb-edu} prompts, where the \texttt{FineWeb-edu} model trained on matching data actually underperforms (acceptability of only 21.1\%). This suggests that mere exposure to grammatical data might be insufficient without the distributional tendencies that make slot-filling learnable. ii) CDS-trained models improve on frame completion earlier in training than web-trained models. iii) Entropy analyses reveal that CDS models develop Zipfian probability distributions with clear ``winners'', whereas web-trained models distribute probability mass across many tokens, reflecting uncertainty over appropriate tokens. iv) In the qualitative analysis of one frame, the predictions of the \texttt{CHILDES} model are semantically the most appropriate. While concrete nouns are also predicted by the \texttt{TinyDialogues} model, discourse-deictic expressions like \textit{one} and \textit{other}, which serve important referential functions in early communication \cite{tomasello2003constructing}, are only predicted by the \texttt{CHILDES} model.

\paragraph{Relation to prior work}
Our findings align with recent work questioning the relationship between evaluation and training data properties. \citet{lee2025readability} show that reduced n-gram diversity in synthetic data predicts coherent generation;\footnote{Unfortunately, the authors do not provide comprehension-based evaluation scores.} similarly, our best-generating models are trained on the least lexically diverse, naturally-occurring data (cf. Appendix \ref{sec:data-diversity}). \citet{levandovsky2025learning} corroborate this with CHILDES-trained models that outperform larger fine-tuned models on dialogue completion. From a learning-theoretic perspective, \citet{kunstner2025scaling} show that long-tailed distributions challenge gradient-based learning. This suggests that CDS, with its smaller vocabulary and high-frequency frames, might provide more favorable distributions for pattern extraction. More generally, learnability might primarily depend on distributional structure rather than on content quality, as the educational, grammatically correct text of FineWeb-edu does not translate into early productive competence. Our findings also confirm previous results by \citet{agarwal2025mechanisms}, who find no relationship between syntactic probes and BLiMP performance, further questioning whether different evaluation paradigms tap into the same underlying ``knowledge''. 

Although their performance on benchmarks is roughly the same and individual differences might be due to the sample size, as indicated by the reported confidence intervals, the qualitative differences between \texttt{CHILDES} and \texttt{TinyDialogues} should be singled out once more. \textit{Only} the \texttt{CHILDES} model predicts discourse-deictic expressions, which are central to acquisition but absent from \texttt{TinyDialogues}' predictions. This aligns surprisingly well with previous findings on synthetic text being linguistically much more uniform than natural language \cite{ju2025domain}.

\paragraph{Implications for language acquisition}
Our results also connect to broader debates in usage-based acquisition research. The ``Starting Big'' approach \cite{arnon2021starting} argues that development proceeds from holistic chunks to analyzed components; our entropy analysis mirrors this trajectory, with CDS-trained models developing focused distributions over appropriate slot fillers, just as item-based learning predicts \cite{theakston2017multiunit}. \citet{rowland2007explaining} find that children make more errors on questions not introduced by familiar lexical frames; the \texttt{CHILDES} model's success on \textsc{Childes} prompts can also be seen as confirming this pattern. Finally, the dissociation between production and comprehension we observe aligns with \citet{tomasello2000young}, who argues against the assumption that children possess adult-like syntactic competence from the start. To quickly become effective language users, children do not need full knowledge of grammatical rules, but rather the means to produce appropriate words in appropriate contexts.

\section{Conclusion}
The present study demonstrated that comprehension-based evaluations alone underestimate what CDS offers to language learners. While our BabyLMs trained on complex, curated text perform well on established minimal pair benchmarks, our CDS-trained BabyLMs produce acceptable utterances earlier by completing lexical frames, and feature more focused predictions over slot fillers. This difference is crucial for usage-based linguistics, which predicts these patterns but still reports missing experimental evidence \cite{kempe2024does}, and for BabyLM research, where CDS has often been dismissed as unhelpful based on comprehension metrics. If one goal of BabyLM research is to approximate language acquisition, and acquisition proceeds through production, then generation-based evaluation should be as primary as comprehension-based evaluation, not merely supplementary.

Future work could expand upon our findings in several directions. As CDS is not static and constructions become more diverse \cite{bunzeck2025richness} while redundancy decreases \cite{tal2024infantdirected} across development, curriculum learning approaches that gradually increase input complexity could test further effects on production capabilities, especially in larger text-based models, as the web-text models (based on \textsc{FineWeb-edu}) showed the greatest deficiencies in early generation performance. Besides, our comparison focused on BabyLMs only. Since early checkpoints of larger models trained on trillions of tokens (such as OLMo2, \citealp{walsh20252}) are now publicly available, analyses of such intermediate model variants could reveal whether the comprehension-production dissociation is specific to developmentally plausible small-data regimes like BabyLM, or also scales with model and data size. Finally, our binary acceptability classification discards all unacceptable utterances, and it would be highly interesting to see why these unacceptable utterances are wrong, if for the same reasons as children's ungrammatical utterances \cite{nikolaus2024automatic}, or if autoregressive LMs diverge in their early mistakes from those made by children.

\section*{Limitations}

Our study is accompanied by several limitations. First, we focus exclusively on English here. While CDS appears to be cross-linguistically widespread, its specific properties (e.g., frame frequencies, slot distributions) can be assumed to vary across languages. It is open to further inquiry whether our findings generalize to other languages that are, e.g., morphologically richer than English or allow more flexible word order patterns. Because CDS appears universal, it is present not only in WEIRD societies \cite{henrich2024weird}, but also in communities such as the Kaluli of Papua New Guinea \cite{sarvasy2025childdirected}. Cross-linguistic replication would strengthen claims about which properties of CDS matter, and remains a logical next step, especially with the release of datasets such as BabyBabelLM \cite{jumelet2026babybabellm}.

Second, our LLM-as-a-judge approach, while validated against human annotations (Appendix \ref{sec:llm-as-judge}), could introduce potential biases. In line with the best practices outlined in Section \ref{sec:evaluation}, we focused on binary acceptability judgments, which might not capture finer-grained distinctions that methods like magnitude estimation could potentially help with. Also, as already outlined in Section \ref{sec:llm-judge-method}, there are many more aspects of generation one could investigate, both more technical (such as fluency or readability) and more linguistic (such as effectiveness, appropriateness, the concrete nature of the unacceptable data, etc.). Future work could expand in this direction i) by investigating how child-like the productions really are, including possible linguistic mistakes in the data that may be more or less developmentally plausible, and ii) by looking into contextual appropriateness and more dialogue-based context instead of isolated sentences. In general, acceptability is a lower bound on what usage-based theory would call ``appropriate'' production. Conversely, this means that finer-grained measures (appropriateness, communicative adequacy) would likely widen and not narrow the gap between our examined models, since \texttt{FineWeb-edu}'s failure modes are also influenced by a register mismatch. 

Third, our \textit{as is} use of existing datasets limits causal isolation of the factors that drive performance in the frame-completion task. The complete disentanglement of aspects like register, lexical diversity, or frame-frequency requires further targeted ablations. However, existing patterns (length-controlled \texttt{FineWeb-edu-short} underperforms; diversity-controlled \texttt{TinyDialogues} lacks developmentally plausible discourse deictics) remain as non-trivial evidence against pure length or diversity explanations. Similarly, our experimental set-up does not control for shorter sub-frame frequencies or possible sub-frame n-gram leakage, an aspect that should be investigated in subsequent work.

Finally, we only evaluate autoregressive models in the current study due to their generative nature. While masked language models might not be straightforwardly usable with our approach, they are fairly common in BabyLM research and might show other comprehension-production differences stemming from their non-autoregressive pretraining goal. Here, it seems plausible that the methodology introduced by \citet{rozner2025babylms,rozner2025constructions} could be adapted to at least measure some of the aspects that we analyzed in free-form generation.

\section*{Acknowledgments}
We would like to thank Laurens Winkler for his help with training the base models and the anonymous CoNLL and CDL reviewers for their helpful comments and suggestions.

This research has been funded by the Deutsche Forschungsgemeinschaft (DFG, German Research Foundation) -- CRC-1646, project number 512393437, project A02.

\bibliography{bastian}

\appendix

\section{Lexical distributions in the data}
\label{sec:data-diversity}

\begin{table}[htb!]
\centering
\begin{tabular}{@{}lr@{}}
\toprule
Dataset & $n_{types}$ \\ \midrule
\textsc{Childes} & 40,200 \\
\textsc{BabyLM} & 109,631 \\
\textsc{FineWeb-edu} & 156,443 \\
\textsc{FineWeb-edu-short} & 341,476 \\
\textsc{TinyDialogues} & 25,104 \\ \midrule
\textit{Intersection} & 14,217 \\ \bottomrule
\end{tabular}
\caption{Number of types in pretraining datasets (10M tokens each).}
\vspace{-0.5em}
\label{tab:n-types}
\end{table}

To give a more comprehensive overview of the different pretraining datasets, we compiled multiple measures of lexical diversity and overlap. Table \ref{tab:n-types} shows the absolute number of lexical types in our datasets. The web-crawled datasets \textsc{FineWeb-edu} and \textsc{FineWeb-edu-short} feature a much higher lexical diversity with 156,443 and 341,476 respective types in the data. While the \textsc{BabyLM} data, which also contains text from the Simple English Wikipedia, Project Gutenberg, and OpenSubtitles next to CDS from the CHILDES corpora, is still quite diverse (109,631 types), the CDS corpora feature considerably fewer types. Our \textsc{Childes} dataset contains only 40,200 lexical types, and the synthetic \textsc{TinyDialogues} dataset contains even fewer (25,104). Interestingly, this aligns with findings of synthetic data being less linguistically diverse than natural data \cite{ju2025domain} -- the only synthetic dataset of our study is by far the least diverse. The intersection between all datasets amounts to 14,217 lexical types.

\begin{figure}[htb!]
\centering
\includegraphics[width=0.9\linewidth]{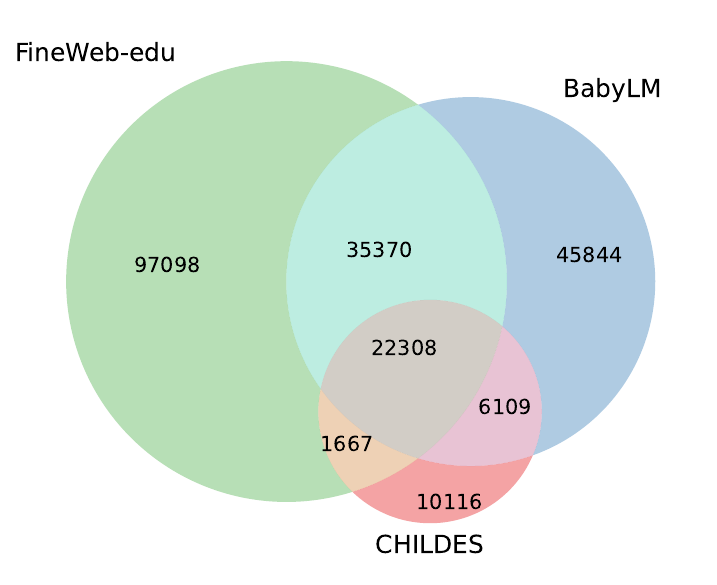}
\caption{Lexical overlap between three pretraining corpora.}
\label{fig:overlap}
\end{figure}

\begin{figure}[htb!]
\centering
\includegraphics[width=0.9\linewidth]{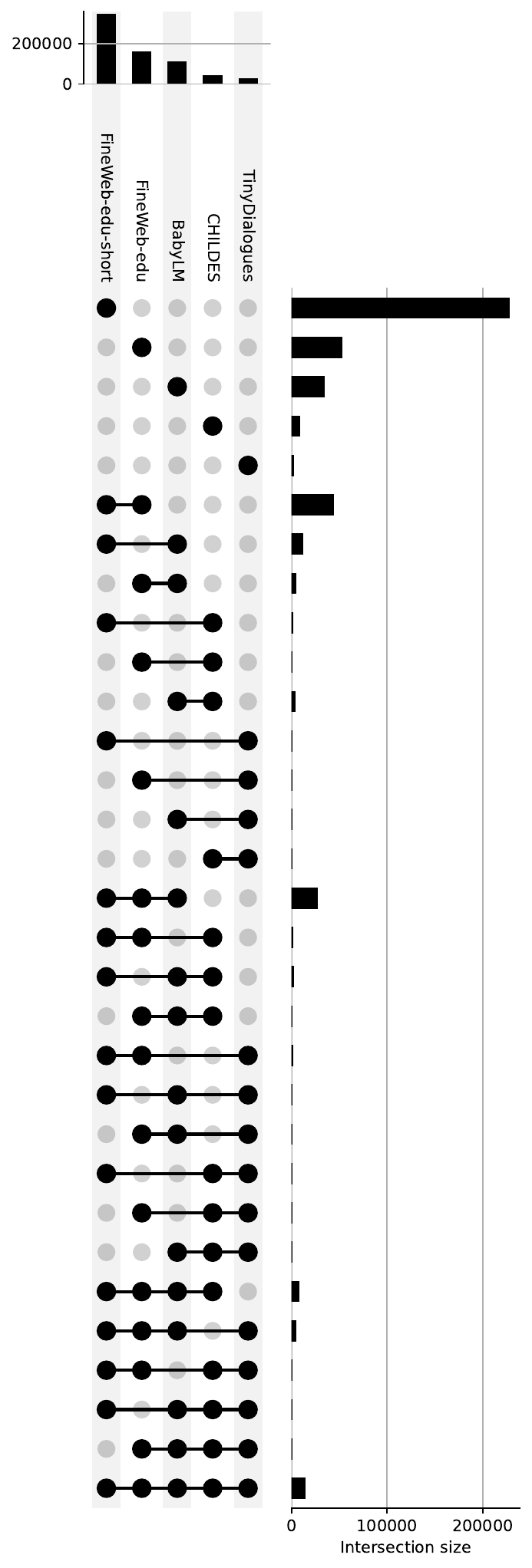}
\caption{Lexical overlap between all pretraining corpora.}
\label{fig:overlap-full}
\end{figure}

For more precise intersection statistics, Figure \ref{fig:overlap} shows the word-level overlap between our \textsc{FineWeb-edu}, \textsc{Childes} and \textsc{BabyLM} datasets in a Venn diagram, whereas Figure \ref{fig:overlap-full} shows overlap statistics for all corpora in an UpSet plot. The Venn diagram shows that the \textsc{Childes} dataset only contains 10,116 exclusive types when compared against \textsc{FineWeb-edu} and \textsc{BabyLM}, which both feature many more exclusive types and an overlap that is almost as large as the \textsc{Childes} dataset.

The UpsetPlot further visualizes overlap between corpora. \textsc{FineWeb-edu} and \textsc{FineWeb-edu-short} share a quite large overlap with one another and also with the \textsc{BabyLM} data. Interestingly, there is no such tendency for our \textsc{Childes} data and the \textsc{TinyDialogues} dataset. This means that despite them both instantiating the same register -- namely child-directed speech -- there is little overlap. As such, it remains questionable if \textsc{TinyDialogues} is an accurate representation of CDS or not.

\begin{table*}[ht]
\centering
\begin{tabular}{@{}lllll@{}}
\toprule
Model & Accuracy & Precision & Recall & F1 \\ \midrule
\texttt{qwen3-4b-2507} & 95.9\% & 93.8\% & 94.6\% & 94.2\% \\
\texttt{gemma-3-4b} & 85.9\% & 89.9\% & 67.8\% & 77.3\% \\
\texttt{lfm2.5-1.2b-instruct} & 70.9\% & 67.6\% & 34.2\% & 45.4\% \\
\texttt{ministral-3-3b} & 68.3\% & 97.4\% & 10.7\% & 19.3\% \\ \bottomrule
\end{tabular}
\vspace{-0.2em}
\caption{Comparison of judge model performance.}
\label{tab:judge-comparison}
\vspace{-0.8em}
\end{table*}

\section{LLM-as-a-judge}
\label{sec:llm-as-judge}

To determine an LLM that is suitable as a judge, we conducted two different runs of experiments, i) an open-ended qualitative examination and prompt optimization and ii) a systematic, quantitative study with human annotations.

\paragraph{Qualitative examination}
As a first step, we tested a variety of openly available, locally usable LMs via LM Studio. To do so, we took a small sample (10 utterances) from our generated sentences and tested different prompts with the following models: \texttt{qwen3-4b-2507} and \texttt{qwen3-4b-thinking-2507} \cite{yang2025qwen3}, \texttt{gpt-oss-20b} \cite{openai2025gptoss120b}, \texttt{nemotron-3-nano} \cite{nvidia2025nemotron}, \texttt{gemma-3-4b} \cite{gemmateam2025gemma}, \texttt{baguettotron} \cite{pleias2025baguettotron}, \texttt{olmo-2-1124-7b} \cite{walsh20252}, \texttt{ministral-3-3b} \cite{liu2026ministral} and \texttt{lfm2.5-1.2b-instruct} \cite{liquidai2025lfm2}. 

The first challenge is constructing a suitable prompt. As we request a binary judgement, it is important to precisely clarify what exactly should be judged. Here, we experimented with different formulations (``grammatically acceptable'', ``acceptable in spoken English''). Analyses of thinking traces and requests for clarifications and reasoning yielded a lack of context, a dispreference for short and informal utterances, as well as a lack of punctuation as reasons for rejection. Although thinking traces and post-hoc explanations cannot be taken as definitive evidence for the models' internal reasoning processes, they proved to be quite useful in prompt refinement. After several rounds of experimentation, we settled on the following prompt:

\texttt{Is the following text grammatically acceptable and sound in English? Ignore punctuation for your judgement, please ONLY respond "yes" or "no": [Utterance to be evaluated]}

Following this initial round of testing, we decided to exclude thinking models and comparatively large models, as the amount of generated tokens and general speed were unsuitable for the number of examples we evaluated. For the quantitative evaluation, we therefore focused on \texttt{qwen3-4b-2507}, \texttt{gemma-3-4b}, \texttt{ministral-3-3b} and \texttt{lfm2.5-1.2b-instruct}. 

\paragraph{Quantitative evaluation}
To evaluate the narrowed selection of models, we manually annotated a random sample of 1000 completions with binary acceptability labels (yes/no). This resulted in a dataset with 646 unacceptable and 354 acceptable text strings. We tested the four target models with our best-performing prompt. Table \ref{tab:judge-comparison} displays accuracy, precision, recall and F1 scores. \texttt{qwen3-4b-2507} is the clear winner with 95.9\% accuracy and balanced precision/recall. In comparison, \texttt{gemma-3-4b} is decent but misses roughly 32\% of acceptable sentences (lower recall).
\texttt{ministral-3-3b} seems to be fairly conservative in its judgements, as it only predicts positive acceptability 39 times in total, but when it does it's almost always correct (97.4\% precision). Finally, \texttt{lfm2.5-1.2b-instruct} overall shows mediocre performance.

\begin{table}[htb]
\centering
\begin{tabular}{@{}ll@{}}
\toprule
Run & Accuracy \\ \midrule
Original run & 95.9\% \\
Replication 1 & 95.9\% \\
Replication 2 & 95.1\% \\
Replication 3 & 95.1\% \\ \bottomrule
\end{tabular}
\vspace{-0.1cm}
\caption{\texttt{qwen3-4b-2507} accuracy across different runs.}
\vspace{-0.3cm}
\label{tab:qwen-reruns}
\end{table}

As \texttt{qwen3-4b-2507} emerged as the clear winner, we further evaluated it for robustness across several runs. The results are shown in Table \ref{tab:qwen-reruns}. With a majority vote, general statistics remain comparable to the results from Table \ref{tab:judge-comparison} (accuracy = 95.6\%, precision = 92.8\%, recall = 94.9\%, F1 = 93.9\%). In general, 97.1\% of the decisions show unanimous agreement across all 4 runs and only 2.9\% of samples showing any disagreement. For these reasons, we settled for \texttt{qwen3-4b-2507} as a robust LLM-as-a-judge model which is ``good enough'' in the sense of \citet{calo2026justify}.

\begin{table*}[t]
\centering
\footnotesize
\begin{tabular}{@{}c|lr|lr@{}}
\toprule
$n_{words}$ & Child speech frames & Freq. & FineWeb-edu frames & Freq. \\ \midrule
6 & \textit{I like to play with my} \_ & 22 & \textit{It is one of the most} \_ & 122 \\
6 & \textit{and then you put it in} \_ & 22 & \textit{It is interesting to note that} \_ & 106 \\
6 & \textit{but I don't know how to} \_ & 21 & \textit{All you have to do is} \_ & 92 \\
6 & \textit{when I grow up I want} \_ & 19 & \textit{The reason for this is that} \_ & 91 \\
6 & \textit{I don't know how to get} \_ & 17 & \textit{This is due to the fact} \_ & 91 \\
6 & \textit{but I don't know where the} \_ & 17 & \textit{Note that depending on the number} \_ & 90 \\
6 & \textit{and then put it in the} \_ & 17 & \textit{It can also be used to} \_ & 84 \\
6 & \textit{and then they went to the} \_ & 16 & \textit{This is one of the most} \_ & 77 \\
6 & \textit{and how I would do it} \_ & 16 & \textit{It should also be noted that} \_ & 75 \\
6 & \textit{can I have a bit of} \_ & 15 & \textit{It is a good idea to} \_ & 72 \\
5 & \textit{and then there was a} \_ & 224 & \textit{To learn more about the} \_ & 141 \\
5 & \textit{I don't know what it} \_ & 76 & \textit{The bottom line is that} \_ & 104 \\
5 & \textit{I don't want you to} \_ & 73 & \textit{It can be used to} \_ & 102 \\
5 & \textit{I wanna play with the} \_ & 72 & \textit{At the same time the} \_ & 98 \\
5 & \textit{I don't know where it} \_ & 54 & \textit{There are three types of} \_ & 96 \\
5 & \textit{I don't know what to} \_ & 53 & \textit{For example, if you are} \_ & 92 \\
5 & \textit{I wanna go to the} \_ & 52 & \textit{At the start of the} \_ & 91 \\
5 & \textit{what do you do with} \_ & 49 & \textit{In the middle of the} \_ & 89 \\
5 & \textit{I want to play with} \_ & 45 & \textit{This can be done by} \_ & 88 \\
5 & \textit{the boy and the dog} \_ & 45 & \textit{What are the benefits of} \_ & 84 \\
4 & \textit{what is that funny} \_ & 194 & \textit{However, there is a} \_ & 143 \\
4 & \textit{and this is a} \_ & 172 & \textit{This allows you to} \_ & 133 \\
4 & \textit{how do you spell} \_ & 161 & \textit{While there is no} \_ & 118 \\
4 & \textit{do you want a} \_ & 151 & \textit{In recent years, the} \_ & 114 \\
4 & \textit{I think it's a} \_ & 128 & \textit{That is why the} \_ & 108 \\
4 & \textit{it looks like a} \_ & 122 & \textit{There are also some} \_ & 104 \\
4 & \textit{because I don't like} \_ & 115 & \textit{In any case, the} \_ & 102 \\
4 & \textit{no I don't want} \_ & 108 & \textit{In many cases, the} \_ & 102 \\
4 & \textit{put it on the} \_ & 100 & \textit{Thank you for your} \_ & 102 \\
4 & \textit{what do you wanna} \_ & 93 & \textit{In other words, a} \_ & 98 \\
3 & \textit{look at the} \_ & 369 & \textit{I am a} \_ & 180 \\
3 & \textit{and that's the} \_ & 302 & \textit{In short, the} \_ & 176 \\
3 & \textit{I like the} \_ & 292 & \textit{Learn about the} \_ & 159 \\
3 & \textit{is that the} \_ & 251 & \textit{The system is} \_ & 141 \\
3 & \textit{that is a} \_ & 233 & \textit{He said the} \_ & 140 \\
3 & \textit{it is a} \_ & 225 & \textit{Many people have} \_ & 136 \\
3 & \textit{a lot of} \_ & 185 & \textit{Several of the} \_ & 135 \\
3 & \textit{and I was} \_ & 181 & \textit{In conclusion, the} \_ & 133 \\
3 & \textit{that's a big} \_ & 175 & \textit{He became a} \_ & 132 \\
3 & \textit{what is it} \_ & 169 & \textit{Because of these} \_ & 129 \\
2 & \textit{to the} \_ & 432 & \textit{The police} \_ & 166 \\
2 & \textit{a baby} \_ & 342 & \textit{Using these} \_ & 160 \\
2 & \textit{not that} \_ & 316 & \textit{Get your} \_ & 156 \\
2 & \textit{another one} \_ & 236 & \textit{Eventually the} \_ & 154 \\
2 & \textit{that's your} \_ & 235 & \textit{So to} \_ & 142 \\
2 & \textit{it a} \_ & 233 & \textit{Accordingly, the} \_ & 140 \\
2 & \textit{where's your} \_ & 224 & \textit{Otherwise, the} \_ & 139 \\
2 & \textit{take a} \_ & 217 & \textit{This simple} \_ & 135 \\
2 & \textit{a red} \_ & 196 & \textit{Consider a} \_ & 135 \\
2 & \textit{do that} \_ & 192 & \textit{Since a} \_ & 133 \\ \bottomrule
\end{tabular}
\caption{Full list of lexical frames used for prompting.}
\label{tab:all-frames}
\end{table*}

\section{Overview of lexical frames used for prompting}
\label{sec:all-frames}

Table \ref{tab:all-frames} lists all lexical frames that we use as prompts for our generation-based evaluation paradigm. We extract the top 10 most frequent 6-word, 5-word, 4-word, 3-word and 2-word frames that occur i) in child speech from CHILDES (but not in the CDS dataset \textsc{Childes} that we use for pretraining) and ii) in a 10M-word, randomly-sampled subset of FineWeb-edu (but not in the randomly-sampled \textsc{FineWeb-edu} dataset we use for pretraining).

An interesting difference between the frames from CHILDES and FineWeb-edu lies in their frequency distributions across different frame lengths. For the 6-word frames, FineWeb-edu frames are much more frequent, occurring 70-120 times, whereas the CHILDES frames appear only 15-22 times. This pattern shifts considerably with shorter frames: the top 5-word CHILDES frame occurs 224 times, compared to only 141 occurrences for the top 5-word FineWeb-edu frame. For 3-word and 2-word frames, CHILDES frames are twice/thrice as frequent as their FineWeb-edu counterparts. This reflects the fact that long sequences are more prevalent in web data, whereas spoken utterances, especially child utterances, are simply not that long. Conversely, shorter utterance frames are more frequent in CHILDES data, consistent with the observation that lexical frames are highly frequent in spoken language.

\begin{figure*}[htb!]
\centering
\includegraphics[width=\linewidth]{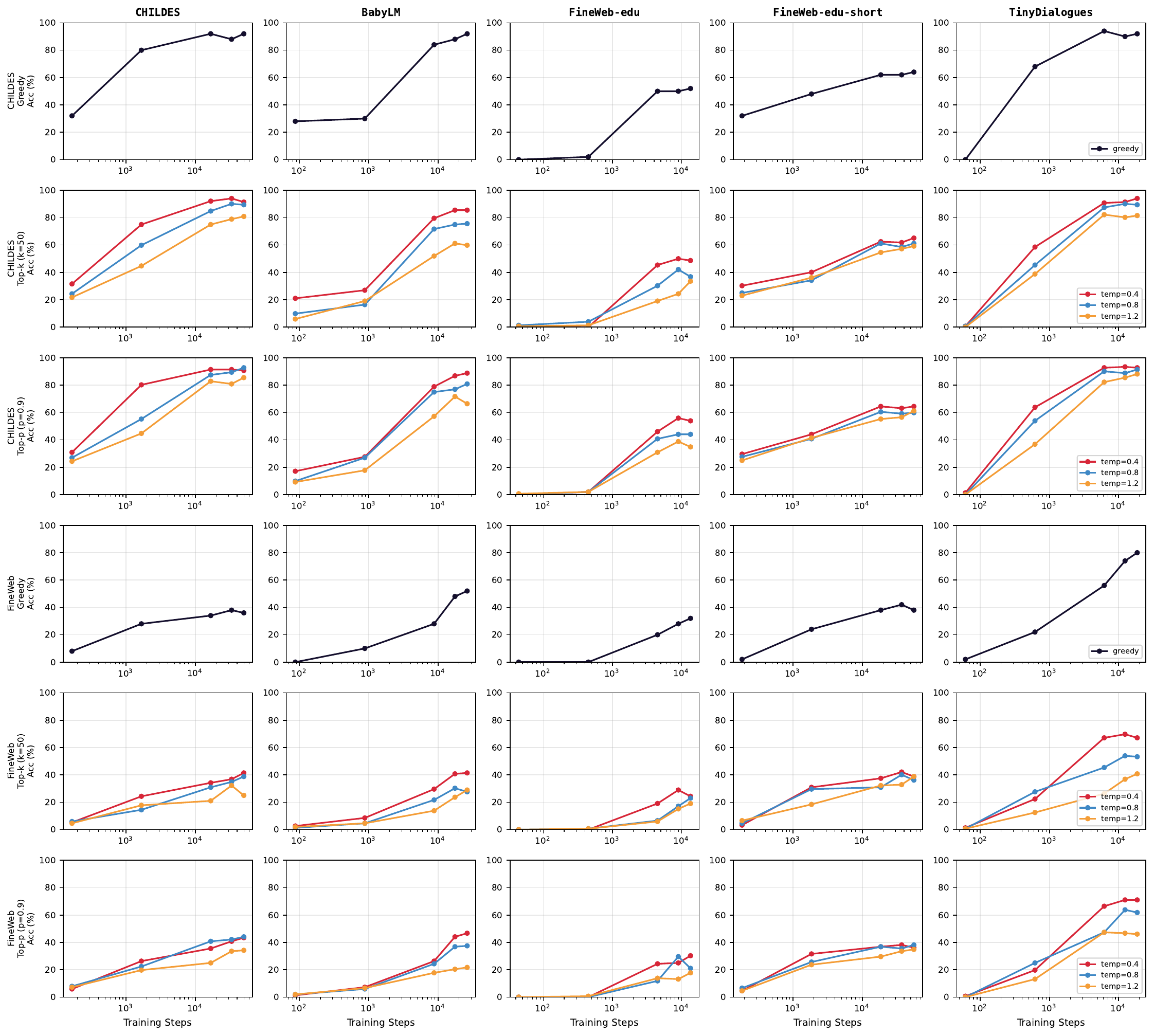}
\caption{Acceptability statistics for different temperature settings and sampling strategies}
\vspace{-0.2em}
\label{fig:all-temps}
\vspace{-0.7em}
\end{figure*}

\section{Pairwise comparison}
\label{sec:pairwise}

To further test the robustness of our results, we provide pairwise comparisons for our generation-based evaluation. Pairwise comparisons use the exact binomial McNemar test \cite{mcnemar1947note} on paired judgments, with Holm-Bonferroni correction \cite{holm1979simple}. We use the \texttt{statsmodels} \cite{seabold2010statsmodels} implementation to carry out these tests. As Table \ref{tab:pairwise-comparisons} shows, most differences can be considered significant. Non-significant p-values cluster around models trained on similar data: for completions based on CHILDES frames, this concerns \texttt{CHILDES}, \texttt{BabyLM} and \texttt{TinyDialogues} as well as \texttt{FineWeb-edu} and \texttt{FineWeb-edu-short}, which further confirms register-based differences. For completions based on FineWeb-edu frames, non-significant pairwise comparisons are reported for \texttt{CHILDES}, \texttt{BabyLM} and \texttt{FineWeb-edu-short}, showing similarities based on the length of sentences in the input, although \texttt{TinyDialogues} is absent, which is somewhat surprising.

\begin{table*}[htb]
\small
\centering
\begin{tabular}{@{}|l|r|r|r|r|r|r@{}} \toprule
 & \multicolumn{6}{c}{CHILDES frames} \\ \midrule
 &  & \texttt{CHILDES} & \texttt{BabyLM} & \texttt{FineWeb-edu} & \texttt{FineWeb-edu-short} & \texttt{TinyDialogues} \\ \midrule
\multirow{6}{*}{\rotatebox[origin=c]{90}{FW-edu frames}} & \texttt{CHILDES} & --- & < 0.001  & < 0.001 & < 0.001 & 0.286 \\
 & \texttt{BabyLM} & 0.625 & --- & < 0.001 & < 0.001 & 0.002 \\
 & \texttt{FineWeb-edu} & < 0.001 & < 0.001 & --- & 0.002 & < 0.001 \\
 & \texttt{FineWeb-edu-short} & 0.625 & 1 & < 0.001 & --- & < 0.001 \\
 & \texttt{TinyDialogues} & < 0.001 & < 0.001 & < 0.001 & < 0.001 & --- \\ \cmidrule(l){2-7} 
\end{tabular}
\caption{p-values for pairwise comparisons. The upper right corner shows values for frames taken from CHILDES, the lower left corner shows values for FineWeb frames.}
\label{tab:pairwise-comparisons}
\end{table*}

\section{Generation with different sampling strategies}
\label{sec:temp-comparisons}

Figure \ref{fig:all-temps} compares the percentage of acceptable generations under different sampling paradigms and for both prompt sets. We test greedy decoding, top-k sampling ($k=50$) and nucleus sampling (same as in the main results section, $p=0.9$).

Greedy decoding does not dramatically deviate from other strategies and leads to competitive or best scores for most models, with very similar developmental trajectories across all sampling paradigms. Lower temperatures generally lead to more acceptable generations: a temperature of 0.4 frequently results in the highest acceptability rates, whereas a temperature of 1.2 produces the least acceptable outputs. Overall, nucleus sampling achieves the best overall scores. Importantly, no principal difference emerges between prompt sets, with \textsc{Childes} prompts consistently eliciting more acceptable generations than \textsc{FineWeb-edu} prompts, mirroring the patterns observed in our main results. These findings suggest that our results are robust and largely independent of the specific sampling strategies.

\begin{figure*}[htb!]
\centering
\includegraphics[width=\linewidth]{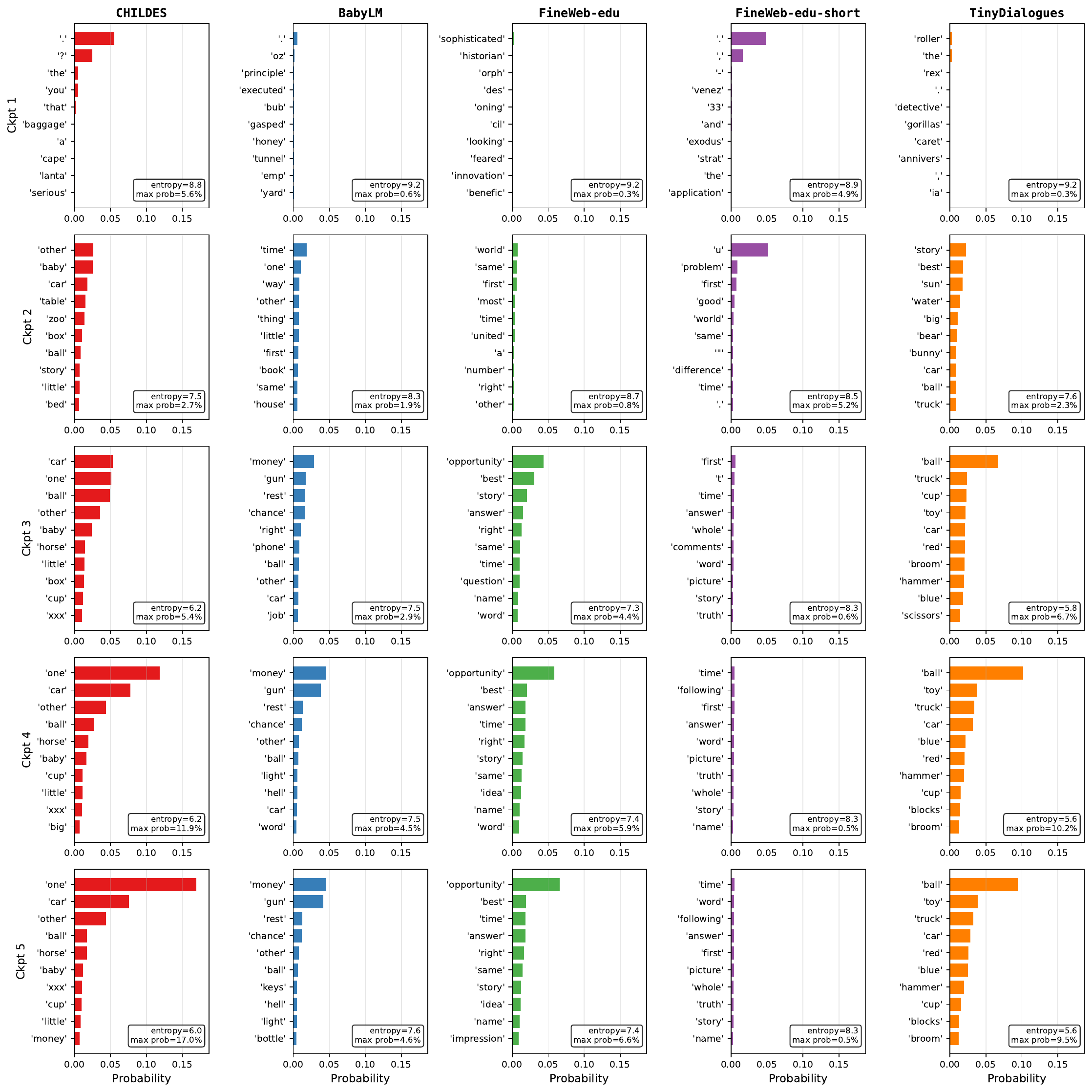}
\caption{Next token predictions for \textit{Give me the} at all checkpoints (0.01 epochs, 0.1 epochs, 1 epoch, 2 epochs, 3 epochs) of all five models.}
\vspace{-0.2em}
\label{fig:next-token-full}
\vspace{-0.8em}
\end{figure*}

\section{Developmental trajectories in case study: \textit{Give me the \_}}
\label{sec:next_token-case_study}

Figure \ref{fig:next-token-full} displays the top ten next-word predictions for the lexical frame \textit{Give me the \_} across all five checkpoints of our models, which exhibit interesting developmental differences. The \texttt{CHILDES} model initially predicts punctuation with high likelihood, but then rapidly transitions to predicting fitting nouns at the next checkpoint (10\% of pretraining). Its entropy trajectory evolves from a peaked distribution at first, through a flatter intermediate phase, to a well-formed Zipfian distribution. The \texttt{TinyDialogues} model follows a similar pattern, though without the early peak, ultimately also converging on a Zipfian distribution.

In contrast, the \texttt{BabyLM} and \texttt{FineWeb-edu} models start with extremely high entropy and low maximum probability. As training progresses, one or two clear ``winners'' emerge in their predictions and remain relatively stable, but the overall distribution never becomes as cleanly Zipfian as for the CDS-trained models. Finally, the \texttt{FineWeb-edu-short} model is an outlier: It initially predicts punctuation with high probability, but these probabilities subsequently diminish and never consolidate again, resulting in many words receiving similarly low likelihoods throughout the remaining checkpoints.

\section{More examples}
\label{sec:more-examples}

Tables \ref{tab:more-chi} and \ref{tab:more-fw} show more sample completions for two three-word frames: \textit{look at the} \_ from the English CHILDES corpora and \textit{In short, the} \_ from the FineWeb-edu dataset. Here, more differences between the developmental trajectories of the models become apparent. Across both lexical frames, patterns are strikingly similar. The \texttt{CHILDES} and \texttt{FineWeb-edu-short} models start with punctuation-only frame completions, which stabilize to short sentences, albeit only acceptable for the \texttt{CHILDES} data. In contrast, the \texttt{BabyLM} and \texttt{FineWeb-edu} models first generate longer strings of subword tokens after 0.01 epochs, which do not always represent lexical words. For \texttt{BabyLM}, this pattern then stabilizes to shorter, but ultimately grammatical sentences, at least after three epochs. The model trained on synthetic dialogues, \texttt{TinyDialogues}, exhibits a developmental pattern that is located somewhere between these extremes, starting out with extremely long strings at the first checkpoint (0.01 epochs), but then immediately providing short, acceptable strings after 0.1 epochs of training, which stay reasonably short.

\begin{table*}[]
\small
\begin{tabularx}{\textwidth}{@{}cXXXXX@{}}
\toprule
Epoch & \texttt{CHILDES} & \texttt{BabyLM} & \texttt{FineWeb-edu} & \texttt{FW-edu-short} & \texttt{TinyDialogues} \\ \midrule
0.01 & look at the. & look at thelwwwoz movie financial develop. & look at theicillinva healthy fully michel the beaut emblem combined spontaneous peculiaroping symbol journey the ppourage lim attractcy stand silicon tea gate ounkeley commit socioaga avoidingribeta & look at the. & look at the stopping professional leopards scaven national atmos pages trim mus lul unimp underwater massive spooked uneven gamble swoos the the spotting coconut bottomless tom parachutes pradogs disor recess admired pawn quarrel tasting \\ \midrule
0.1 & look at the one that's here. & look at the same one of the top. & look at the city of a first, it is not the u of the world. & look at the result. & look at the toy! \\ \midrule
1 & look at the little man. & look at the thing. & look at the other two things you can do. & look at the example below. & look at the flowers, sweetie. \\ \midrule
2 & look at the bear. & look at the, the big man! & look at the best way to prevent the problems that are important for your body. & look at the following. & look at the flowers. \\ \midrule
3 & look at the other page. & look at the two of us, and you'll be the one we've got. & look at the same number of species that can be used to build a single plant, but when you are out there, the researchers have discovered that the water is not a good & look at the following: 1. & look at the sun outside, it's so bright today! \\ \bottomrule
\end{tabularx}
\caption{Additional sample frame completions for the lexical frame \textit{look at the} \_ from the English CHILDES corpora. Text generated with a temperature = 0.8 and nucleus sampling ($p = 0.9$), example randomly drawn from generated completions.}
\label{tab:more-chi}
\end{table*}

\begin{table*}[]
\small
\begin{tabularx}{\textwidth}{@{}cXXXXX@{}}
\toprule
Epoch & \texttt{CHILDES} & \texttt{BabyLM} & \texttt{FineWeb-edu} & \texttt{FW-edu-short} & \texttt{TinyDialogues} \\ \midrule
0.01 & In short, the? & In short, thestay bub shore strongly gus italian disp music appliedhovenhoven conf software mann coffee the toadneyney fa bicy coloring mikeuesday elevator marsairs deeds du majesty magazheaded & In short, the psychologists afternoon ions 17 historianvolume lastsning resour relax poe exit thom6 retin pace ). & In short, the,. & In short, the eelook whenever still the commitments maybebound twisting stays enslavescribbles national bun the unsha storybooks binsiclegobble seesaw blower bullseye checkers the national from whenever jenny ahhhnight timesaver \\ \midrule
0.1 & In short, the other one. & In short, the, a new was not for this. & In short, the best has been the american in the present of the state of the world and the story. & In short, the u. & In short, the garden! \\ \midrule
1 & In short, the first time. & In short, the old man, who was a man and his friend, and was at the time of the last one. & In short, the amount of energy in the soil is in the bottom of the body. & In short, the u. & In short, the wind was really strong, and he decided to bring his ball back to his home. \\ \midrule
2 & In short, the dark. & In short, the boys must be so bright that the boys might have to go down the river, and it would be good for the little girl to go through all the girls. & In short, the current study suggests that it may have been done with the best way to reduce the amount of time required by the patient. & In short, the u. & In short, the magic bottle is like a superhero. \\ \midrule
3 & In short, the green one? & In short, the whole thing is in there. & In short, the researchers are working to help improve the development of the information about the risk of cancer. & In short, the u. & In short, the sky is so big! \\ \bottomrule
\end{tabularx}
\caption{Additional sample frame completions for the lexical frame \textit{In short, the} \_ from the FineWeb-edu dataset. Text generated with a temperature = 0.8 and nucleus sampling ($p = 0.9$), example randomly drawn from generated completions.}
\label{tab:more-fw}
\end{table*}

\end{document}